\documentclass[10pt,a4paper]{IEEEtran} 
\usepackage[a4paper,bindingoffset=0.2in,%
            left=0.6in,right=0.6in,top=0.8in,bottom=0.8in,
            footskip=.25in]{geometry}
\renewcommand\IEEEkeywordsname{Keywords}
\usepackage{amsmath,graphicx}
\usepackage{color, colortbl, epstopdf, multirow, makecell}
\usepackage{subfigure}
\usepackage{tabulary}
\usepackage{threeparttable}
\usepackage{cite}
\usepackage{nopageno}
\usepackage{amsmath}
\DeclareMathOperator*{\argmax}{argmax}

\usepackage[pscoord]{eso-pic}

\begin{document}
    
\title{Two-stream Fusion Model for Dynamic Hand Gesture Recognition using 3D-CNN and 2D-CNN Optical Flow guided Motion Template}

\author{\IEEEauthorblockN{Debajit Sarma, V. Kavyasree and M.K. Bhuyan}\\
\IEEEauthorblockA{ Department of Electronics and Electrical Engineering, IIT Guwahati, India 781039\\
Email: \textit{\{s.debajit, kavyasre, mkb\}}\emph{@}\textit{iitg.ac.in}}\vspace{-1.9em} }
  
\maketitle
\thispagestyle{empty}
\begin{abstract}
The use of hand gestures can be a useful tool for many applications in the human-computer interaction community. In a broad range of areas hand gesture techniques can be applied specifically in sign language recognition, robotic surgery, augmented and virtual reality, \textit{etc}. In the process of hand gesture recognition, proper detection, and tracking of the moving hand become challenging due to the varied shape and size of the hand. Here the objective is to track the movement of the hand irrespective of the shape, size and color of the hand. And, for this, a motion template guided by optical flow (OFMT) is proposed. OFMT is a compact representation of the motion information of a gesture encoded into a single image. In the experimentation, different datasets using bare hand with an open palm, and folded palm wearing green-glove are used, and in both cases, we could generate the OFMT images with equal precision. Recently, deep network-based techniques have shown impressive improvements as compared to conventional hand-crafted feature-based techniques. Moreover, in the literature, it is seen that the use of different streams with informative input data helps to increase the performance in the recognition accuracy. This work basically proposes- a two-stream fusion model for hand gesture recognition and a compact yet efficient motion template based on optical flow. Specifically, the two-stream network consists of two layers- a 3D convolutional neural network (C3D) that takes gesture videos as input and a 2D-CNN that takes OFMT images as input. C3D has shown its efficiency in capturing spatio-temporal information of a video. Whereas OFMT helps to eliminate irrelevant gestures providing additional motion information. Though each stream can work independently, they are combined with a fusion scheme to boost the recognition results. We have shown the efficiency of the proposed two-stream network on two databases.

\end{abstract}


\begin{figure*}[h]
\center
\includegraphics[height=0.32\linewidth, width=0.97\linewidth]{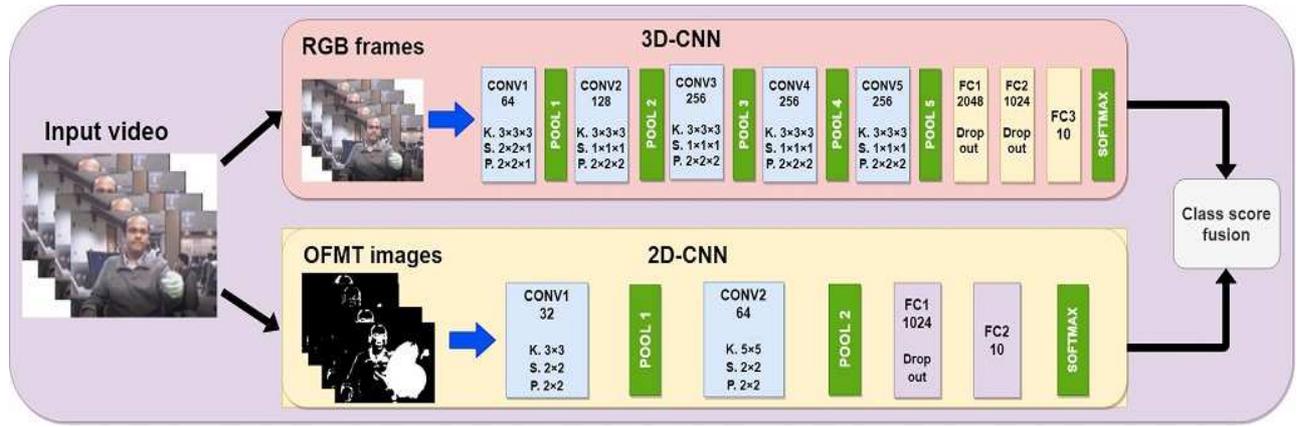}
\caption{Proposed framework for hand gesture recognition (K=kernel size, S=stride size, P=pooling size, max-pooling is used here)}
\label{Fig:Model_diagram}
\end{figure*}
\section{Introduction}

In recent times, a significant effort has been devoted to body/body-part motion analysis in action and gesture recognition.  And, gesture recognition is an important research area in computer-vision with many applications. Since gestures constitute a common and natural means for non-verbal communication, hand gesture recognition from visual images constitute an important part of this research \cite{2006_Karam}. The applications of hand gesture recognition cover various domains, ranging from sign language to medical assistance to virtual reality \cite{2017_Chakraborty}. 


Estimation of the motion field is invariant to shape and appearance (at least in theory) and can be used directly to describe human gesture/action. Optical flow and motion-templates are the two main motion-based representation methods used for this purpose \cite{2019_Sarma}. Generally, both these methods are used separately for motion estimation since both have their own advantages. Motion templates like motion-energy-image (MEI) and motion-history-image (MHI) \cite{2001_Bobick} give a global aspect of motion without the requirement of segmentation of the moving object. It is computationally very efficient making it suitable for real-time applications \cite{2012_Ahad}. In \cite{2013_Mahbub}, authors applied an approach combining MHI with statistical measures and frequency domain transformation on depth images for one-shot-learning hand gesture recognition. Due to the availability of the depth information, the background-subtracted silhouette images were obtained using a simple mask threshold. Whereas in\cite{2019_Zhang}, authors used pseudo-color based MHI images as input to convolutional networks. On the other hand, optical flow is obtained from the movement of the target object in a video scene. Though it is computationally a little expensive, still it has the advantage that it can produce good results even in the presence of a bit of camera movement. In \cite{2011_Mahbub}, optical flow was used to detect the direction of motion along with the RANSAC algorithm which in turn helped to further localize the motion points. There are only a few examples like \cite{2018_Xu_movement} where the optical flow is combined with the motion template. In \cite{2018_Xu_movement}, authors claimed that the combined technique can give a better discrimination power to describe local motions in a global time-space representation. In this work, we have also proposed a motion template driven by optical flow. Our method is different from \cite{2018_Xu_movement} in reducing background noises through an update rule. In our work also, better discrimination can be seen for optical flow guided motion template (OFMT) over conventional motion templates. This combined method can accurately detect the location and thus provide the contour of the moving object just like a tracker. The effectiveness of this method is quite impressive for long and varied video sequences.

On the other hand, feature extraction is one of the most important steps for proper recognition of the action/gesture. Most hand-crafted features usually demand the user to have some prior knowledge and some pre-processing steps. Generally, the feature extraction process needs to segment the body/body-part from the background in the video sequences. For a complex and changing background environment, segmentation may be very difficult due to the variation of shape and appearance of body/body-part depending on many factors like clothing, illumination variation, image resolution, \textit{etc.} In \cite{2018_Sarma}, the authors used the skin segmentation method to segment the hand portion from the background. But this method had issues when there were some skin-color like objects present in the background. So, one major objective in this work is to skip the segmentation part or to adopt some other method for this purpose. This has motivated the development of learning robust and effective representations directly from raw data and deep learning provides a plausible way of automatically learning multiple level features. Indifference to hand-crafted features, there is a growing trend towards feature representations learned by deep neural networks \cite{2018_Khong, 2015_Molchanov, 2018_Sarma, 2019_Zhang}. But, in deep learning techniques, the main requirement is lots of database samples. Several authors have emphasized the importance of using many diverse training examples for CNN's \cite{2012_Krizhevsky}. For datasets with limited diversity, they have proposed data augmentation strategies to prevent CNN from overfitting. In order to avoid overfitting, Molchanov \textit{et al.} \cite{2015_Molchanov} introduced several space-time video augmentation techniques and applied the whole hand gesture video sequences to a 3D-CNN that can extracts features from both spatial and temporal dimensions by performing 3D convolutions \cite{2015_Tran}.


\vspace{-0.09cm}
Ciregan \textit{et al.} \cite{2012_Ciregan} has shown that the use of multi-column deep CNNs with multiple parallel networks to improve recognition rates of single networks by 30-80\% for various image classification tasks. Similarly, for large scale video classification, Karpathy \textit{et al.} \cite{2014_Karpathy} have shown the best results on combining CNNs trained with two separate streams of the original and spatially cropped video frames. Simonyan and Zisserman \cite{2014_Simonyan} proposed separate CNNs for the spatial and temporal streams that are late-fused and that explicitly use optical flow in the context of action recognition. To recognize sign language gestures, Neverova \textit{et al.} \cite{2015_Neverova} employed CNNs to combine color and depth data from hand regions and upper-body skeletons. A two-stream model with two C3D layers that takes RGB and optical flow computed from the RGB stream as inputs were used by \cite{2018_Khong} for action recognition. \cite{2018_Zhu} used a hidden two-stream CNN model which takes only raw video frames as input and directly predicts action classes without explicitly computing optical flow. Here the network predicts the motion information from consecutive frames through a temporal stream CNN that makes the network $10x$ faster \cite{2018_Zhu}, without computing optical flow which is time-consuming. But still, two hidden layers in one stream are computationally not so efficient. Moreover, state-of-the-art performance is achieved through traditional optical flow precomputed for the convolution layer in a two-stream network for action and hand gesture recognition \cite{2014_Simonyan, 2018_Khong, 2019_Zhang}. But this approach of precomputation of optical flow motion vectors through CNN is computationally expensive and storage inefficient \cite{2018_Zhu}. 

So in this paper, our main intention is to propose a resource-efficient network in terms of data and processing power as much as possible without compromising much in its performance. Here we propose a deep learning-based two-stream network as shown in Fig. (\ref{Fig:Model_diagram}). The first layer/stream is a 3D-CNN (C3D) network in the two-stream architecture that captures the spatial as well as temporal information from gesture videos. The second layer is a 2D-CNN model where the input is an optical flow guided motion template (OFMT) image. OFMT is a hybrid representation, proposed in this work that is obtained by combining optical flow with the motion template to get the advantage of both the methods for temporal evaluation analysis. OFMT is used to provide additional motion pattern information which in turn helps to eliminate irrelevant gestures. The output score of both the layers is fused using an ensemble method to boost the final output. The main contributions of our proposed model are as follows: 

\vspace{-0.09cm}
\begin{enumerate}
\item Ground truth flow is required in supervised training for optical flow estimation. But, generally, the ground truth flow is not available except for limited synthetic data \cite{2018_Zhu}. Moreover, computation of optical flow and then learning the mapping from optical flow to action labels is time-consuming as well as storage demanding. So, we have proposed optical flow guided motion template (OFMT) images as input to the 2D-CNN stream which provides additional temporal information in a resource constraint environment.

\item Our method is efficient in terms of computation and storage point of view as we do not need to store the precomputed optical flow. Moreover, the requirement of segmenting the hand portion from the body is not needed and also, the size, shape, and color of the hand have no effect on the OFMT images.

\item A late-fusion scheme is proposed to leverage the
information containing in both RGB gesture videos and motion template modalities. The advantage of the proposed method is that different deep models can provide complementary motion information. The first layer can capture the spatio-temporal information thorough the 3D deep network, while the motion-patterns are obtained using 2D-CNN through OFMT images.
\end{enumerate}
\vspace{-0.7cm}
\section{The Proposed Methodology}
In this section, the framework of the proposed gesture recognition method is introduced in detail.
\vspace{-0.2cm}
\subsection{Proposed framework}
As shown in Fig. (\ref{Fig:Model_diagram}), the proposed gesture model is composed of two main streams/layers- the first layer is a 3D-CNN (C3D) network in a two-stream architecture to capture spatial as well as temporal information of a gesture. The second layer is a 2D-CNN network, where the input is an optical flow guided motion template (OFMT) image. OFMT is a hybrid, compact and robust motion representation, proposed in this work. The OFMT template is obtained by combining optical flow information with the motion template. In this way, we get the advantage of both the methods for temporal evaluation analysis. The OFMT is used to provide motion pattern information, which in turn eliminates irrelevant gestures. The proposed OFMT can nominally reduce computational complexity and  memory requirement. The output of a CNN classifier is a class-membership probability for each of the gestures under consideration, and thus the prediction results of 3D-CNN and 2D-CNN networks are fused through a simple probability-based ensemble method at the decision level to boost the final output by taking advantage of both the models.
 
\subsection{Spatio-temporal feature learning through a 3D convolutional (C3D) model}
The original C3D \cite{2015_Tran} is designed for RGB videos. The number of parameters of the networks depends on the resolution of input frames. The original C3D was trained on the large-scale dataset Sport1M \cite{2014_Karpathy}, which consists of 1.1M videos downloaded from YouTube consisting of 487 sports classes. 2D-CNN is extended to a 3D-CNN by incorporating the temporal dimension of a video sequence. In 2D-CNNs, the dimension of each feature map is $c \times h \times w$, where $c$ represents the number of filters in the convolutional (conv) layer, $h$ and $w$ represents the height and width of the feature map. In 3D-CNNs, the dimension of each feature map is $c\times l \times h \times w$, where additional parameter $l$ represents the number of frames. This network extracts the features which are compact and generic while being discriminative. As we worked on two smaller databases, a slightly different architecture with 5 conv layers is employed which has a smaller number of parameters compared to the original C3D \cite{2015_Tran} with 8 conv layers. The proposed network has 5 space-time conv layers with 64, 128, 256, 256, 256 kernels. Each conv layer is followed by a rectified linear unit (ReLU) and a space-time max-pooling layer. All 3D convolution kernels are of size $3\times3\times3$, that gives the best performance \cite{2015_Tran} with stride $1\times1\times1$. Max pooling kernels are of size $2\times2\times2$ except for the first, where it is $2\times2\times1$ and stride is $2\times2\times1$. The conv layers are followed by two dense layers with 2048 and 1024 neurons and ReLU as the activation function. To avoid over-fitting while learning, there is a dropout in each dense layers. The parameter of dropout is set to 0.4, which means the layer randomly excludes 40\% of neurons. The final dense layer of the classifier has 10 neurons giving us the respective class labels where softmax function is used for activation.

\subsection{2D motion template CNN model}
As illustrated in Fig. (\ref{Fig:Model_diagram}), the proposed 2D motion template CNN model consists of two major parts- motion templates and a 2D-CNN model. The generation of the motion templates is explained in the next section. The 2D-CNN architecture used in our method is a simple structure based on LeNet \cite{1998_Lecun} (shown in Fig. (\ref{Fig:Model_diagram})). The network has 2 conv layers with 32 and 64 kernels followed by 2 fully connected layers of size 1024 and 10. The final dense layer of the classifier has 10 neurons giving us the respective class labels. The size of the kernels is $3\times3$ and $5\times5$ respectively for the two conv layers. Each conv layer is followed by ReLU and $2\times2$ box non-overlapping max-pooling layers with stride 2 in both horizontal and vertical directions. A dropout of 40\% is used in the dense layer with 1024 neurons to avoid over-fitting.

\subsection{Proposed Optical Flow guided Motion Templates (OFMT):}
The input to the 2D-CNN is a compact motion template. For this, a hybrid representation is proposed for encoding temporal information of a gesture by combining optical flow motion information with motion templates. This representation takes advantage of both optical flow and motion-energy-image (MEI) and motion-history-image (MHI) templates. Optical flow indicates the change in image velocity of a point moving in the scene, also called a motion field. Here the goal is to estimate the motion field (velocity vector) which can be computed from horizontal and vertical flow fields. MEI represents where motion has occurred in an image sequence; whereas MHI represents how an object is moving \cite{2012_Ahad}. MEI describes the motion-shape and spatial distribution of motion, and MHI is the function of the intensity of motion of each pixel at that location. MEI-MHI can be implemented by the following algorithm.

\vspace{0.1cm}
{\bf MEI-MHI Algorithm \cite{2001_Bobick}:}
\begin{itemize}
\item {\bf Image sequences}
\begin{equation} \label{eq_16}
I(x,y,t) = (I_1,I_2,...,I_n)   .
\end{equation}
\item {\bf Image binarization} 
\begin{equation} \label{eq_17}
B(x,y,t) = |I(x,y,t)-I(x,y,t-1)|   .
\end{equation}

where, $B(x,y,t) = \left\{\begin{array}{rcl}
1 & ~~~~ \mbox{if} ~~ B(x,y,t)> \xi \\ 0 & \mbox{otherwise}
\end{array}\right.$

\item {\bf MEI}
\begin{equation} \label{eq_18}
E_\tau(x,y,t) = {\bigcup}_{i=0}^{\tau-1} B(x,y,t-i)   .
\end{equation}

\item {\bf MHI} 
\begin{equation*} \label{eq_19}
H_\tau(x,y,t) = \left\{\begin{array}{lcl} \tau & \mbox{if}~~ B(x,y,t)=1 \\ max (0, H_\tau(x,y,t-1)-\delta & \mbox{otherwise}
\end{array}\right.
\end{equation*}
\end{itemize}
\noindent where $\tau$ decides the temporal extent of the motion (in terms of frames) and $\delta$ is the decay parameter.
%

\begin{figure}[t]
\center
\includegraphics[height=0.23\linewidth,width=0.97\linewidth]{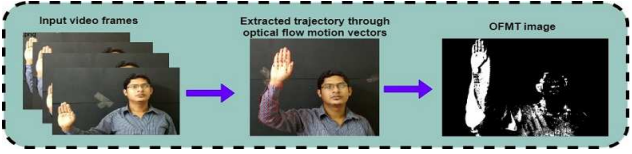}
\caption{Steps to obtain OFMT images}
\vspace{-0.4cm}
\label{Fig:optical_flow_steps}
\end{figure}

\begin{figure}[t]
     \centering
     \subfigure[]{\includegraphics[height=0.12\linewidth, width=0.16\linewidth]{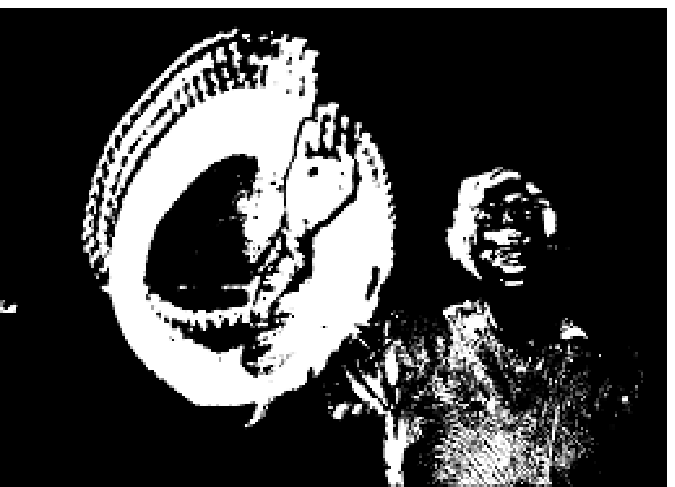}}\quad
     \subfigure[]{\includegraphics[height=0.12\linewidth, width=0.16\linewidth]{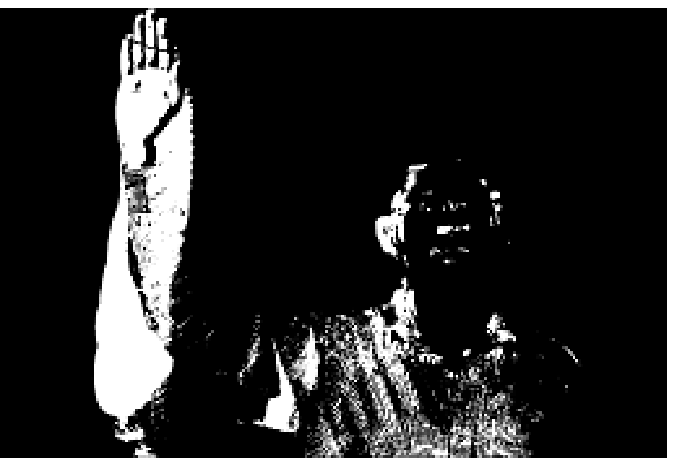}}\quad
      \subfigure[]{\includegraphics[height=0.12\linewidth, width=0.16\linewidth]{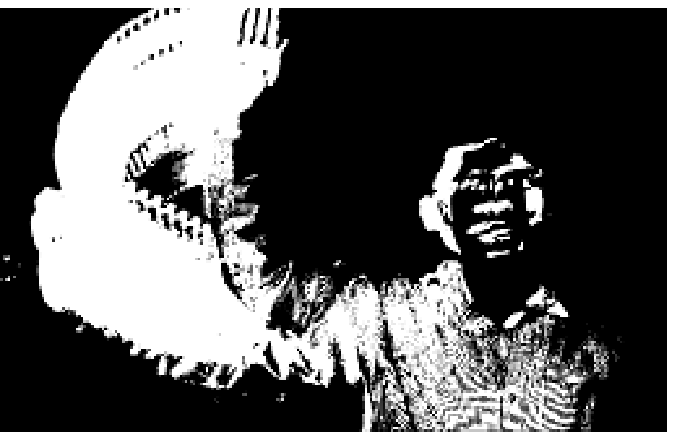}}\quad
	 \subfigure[]{\includegraphics[height=0.12\linewidth, width=0.16\linewidth]{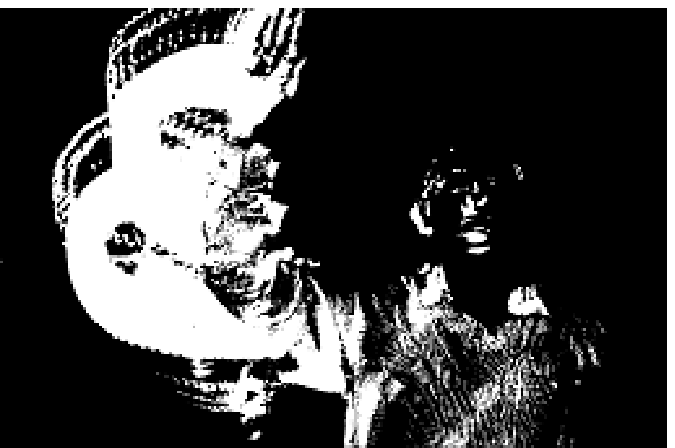}} \quad
	 \subfigure[]{\includegraphics[height=0.12\linewidth, width=0.16\linewidth]{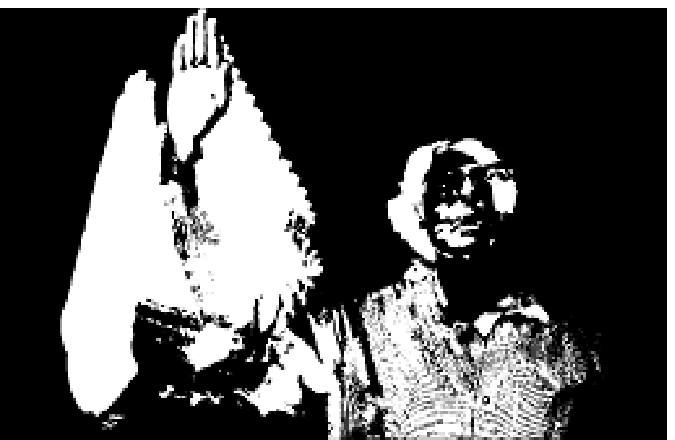}}	
	  
\medskip	 	 

     \subfigure[]{\includegraphics[height=0.12\linewidth, width=0.16\linewidth]{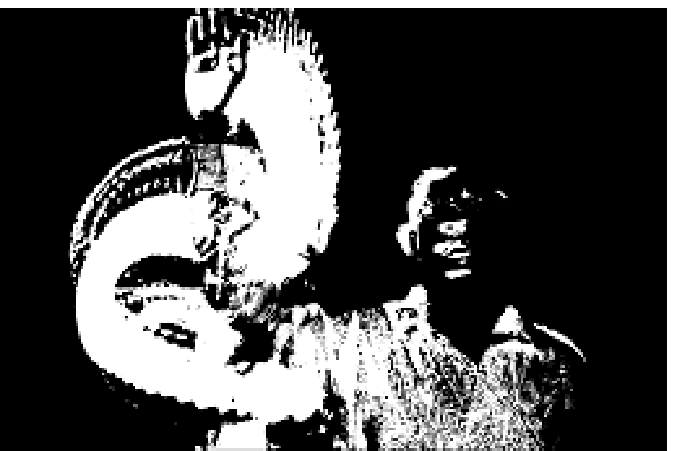}}\quad
     \subfigure[]{\includegraphics[height=0.12\linewidth, width=0.16\linewidth]{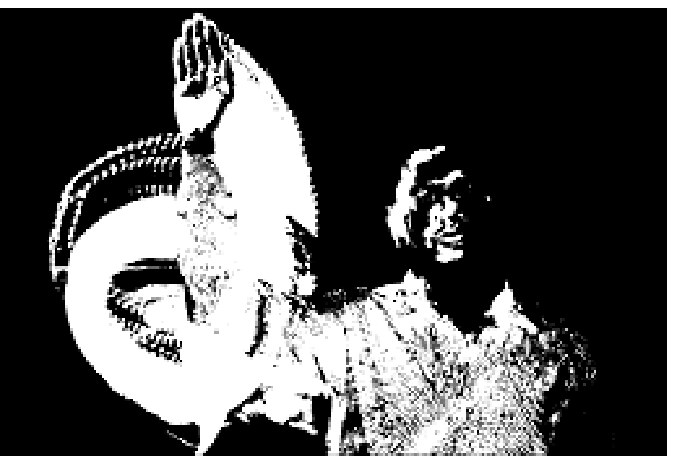}}\quad
	 \subfigure[]{\includegraphics[height=0.12\linewidth, width=0.16\linewidth]{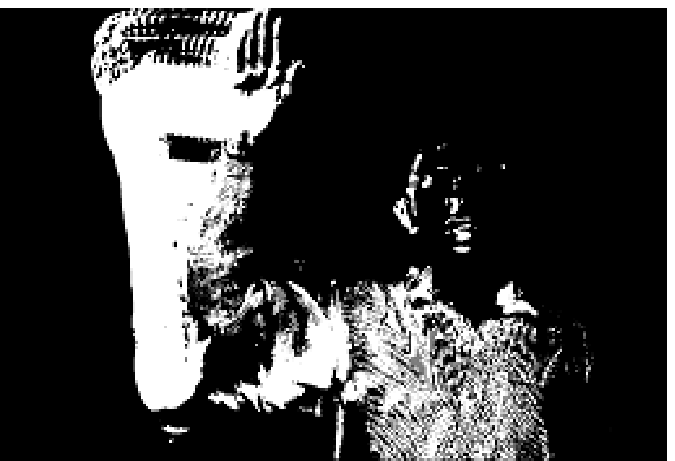}} \quad
     \subfigure[]{\includegraphics[height=0.12\linewidth, width=0.16\linewidth]{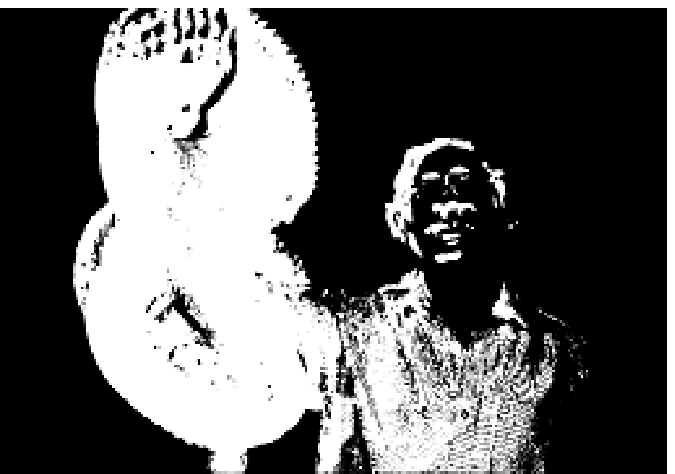}} \quad
	 \subfigure[]{\includegraphics[height=0.12\linewidth, width=0.16\linewidth]{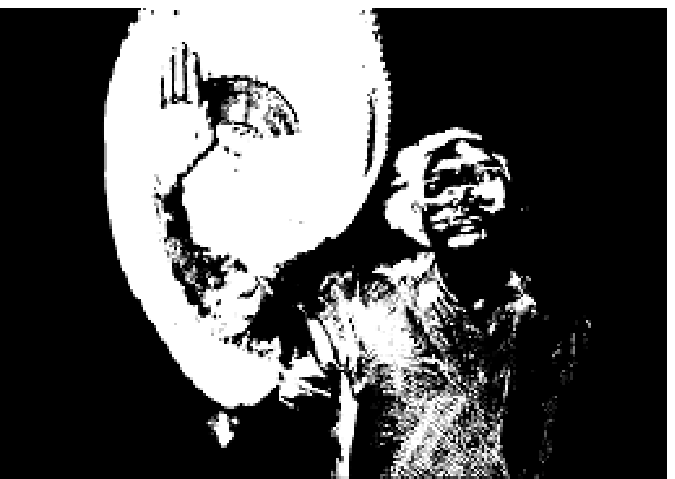}}
	 	
     \caption{Optical Flow guided Motion Templates (OFMT) images applied on our in-house dataset: (a)-(j) for gesture 0-9 }
     \label{Fig:OF_MT_images}
\end{figure}

In this work, we present a motion template driven by Lucas-Kanade \cite{1981_Lucas} optical flow method. This combined method can accurately detect the location and thus provide the contour of the moving object just like an object tracker. MEI and MHI are generated using a binarized image, obtained from frame subtraction, using a threshold $\xi$ as shown in Eq. (\ref{eq_17}). In the motion template representation, all the foreground or moving pixels (\textit{i.e.} $B(x,y,t)=1$) are considered for creating the templates irrespective of the duration and speed of the individual moving pixel. Motion templates basically describe the global motion of a scene and cannot fully describe the local motions of the target object. Whereas optical flow is generally used for foreground segmentation or to extract moving objects. It can better describe the local motions of the target object. In our method, optical flow is judiciously combined  with the motion templates to exploit the advantages of both methods. In the proposed method, the optical flow sequence $O(x,y,t)$ representing the moving regions of the previously smoothed image is accumulated and fused together to form an optical flow guided motion template as:
\begin{equation} \label{eq_20}
E_\tau(x,y,t) = {\bigcup}_{i=0}^{\tau} O(x,y,t-i-1) + \lambda. O(x,y,t)
\end{equation}

\noindent where, $\tau$ indicates the duration of the gesture, and $\lambda$ is an update parameter. If the optical flow length $O(x,y,t)$ is small compared to a pre-defined threshold $\epsilon_s$, then it labels the pixel $(x,y)$ as a background point and hence $\lambda$ value is taken as zero to reduce the effect of background noises. If the optical flow length value is greater than the threshold then it labels a pixel as a foreground moving point, then $\lambda$ is empirically set to 5 to consider foreground pixels. In this way, the background noise is reduced in our proposed method. The saved moving points from the video frames generate a single image providing the trajectory of the gesture as shown in Fig. (\ref{Fig:optical_flow_steps}). These steps are performed for all the hand gesture videos and the corresponding OFMT images obtained from our in-house dataset \cite{2018_Sarma} are shown in Fig. (\ref{Fig:OF_MT_images}). The entire experiment
is done in a python environment, with an OpenCV tool and 30 frames/s is the frame rate used for generating the OFMT images that are fed to the 2D-CNN network. Another advantage obtained here is that the requirement of segmenting the hand portion from the body is not needed and also, the size, shape, and color of the hand have no effect on the OFMT images.




\subsection{Fusion Rules}

In \cite{2014_Simonyan}, the authors used two decision level fusion methods of averaging and SVM fusion, on two identical C3D networks. In the averaging method, two softmax prediction scores are averaged to represent the output class scores. In the SVM fusion method, the features from fully-connected layers of both the streams are stacked and after L2 normalization, are input to an SVM classifier. Whereas in our two-stream model, we have used two non-identical networks applied on inputs with different dimensions. Hence, decision level fusion is preferred here in place of feature-level fusion due to computational overhead. But, in place of just simple averaging fusion, we have adopted an empirical formula given by Eq. (\ref{eq_20}) for output prediction score fusion. 

\begin{equation} \label{eq_20}
p_i = \gamma. p_{i}^{3D} + \delta. p_{i}^{2D} ~~~where~~ i=1,2,...,N
\end{equation}

\noindent Here $p_{i}^{3D}$ and $p_{i}^{2D}$ are the prediction class scores of 3D-CNN and 2D-CNN respectively, $N$ is the number of gesture classes, and $\gamma$ and $\delta$ are two empirical parameters. For fusion at the decision level, different combinations of parameters $\gamma$ and $\delta$ are investigated. By comparing six groups of ($\gamma$, $\delta$), those are (0.8, 0.2), (0.7, 0.3), (0.6, 0.4), (0.4, 0.6), (0.3, 0.7) and (0.2, 0.8), the combination of (0.6, 0.4) for $\gamma$ and $\delta$ achieves the best performance. This is quite justified since the score given by 3D-CNN has a greater impact than the score given by 2D-CNN. The final prediction class score $S$ is the one whose value is maximum and it is calculated as given below:

\begin{equation} \label{eq_21}
S = \argmax\limits_{1 \leq i \leq N} p_i ~~~where~~ i=1,2,...,N
\end{equation}   
\section{Experimentation and Results}
To evaluate the performance of the proposed method, we have carried out experiments on two databases: 1) Palm's Graffiti Digits \cite{2009_Alon} and 2) Our in-house database \cite{2018_Sarma}. The following sections elaborate on all the details during the implementation and evaluation processes.

\subsection{Databases}
In our work, two datasets are employed, one is Palm's Graffiti Digits \cite{2009_Alon} and another is the self-collected in-house dataset. The details are described below.

\begin{enumerate}
\item {\bf Palm's Graffiti Digits}: The Palm's Graffiti digits database \cite{2009_Alon} contains standard RGB 2D videos of ten subjects writing “in the air” the ten Hindu-Western Arabic numerals, 0-9 shown in Fig. (\ref{Fig:Graffiti}), in a continuous streaming mode with video size 320$\times$240, 30 frames/s. This database is split into three subsets, namely “GreenDigits,” “EasyDigits,” and “HardDigits” sets. For our experiments, we have used GreenDigits and EasyDigits. Each one of these two datasets contains 300 gestures in total (ten subjects $\times$ ten digits $\times$ three examples/digit/subject). In GreenDigits set, subjects wear a green glove, while short-sleeves in EasyDigits. In this dataset, the subjects used tightly folded palms while performing the gestures. 

\begin{figure}[htbp]
\center
\includegraphics[height=1.5cm,width=8.5cm]{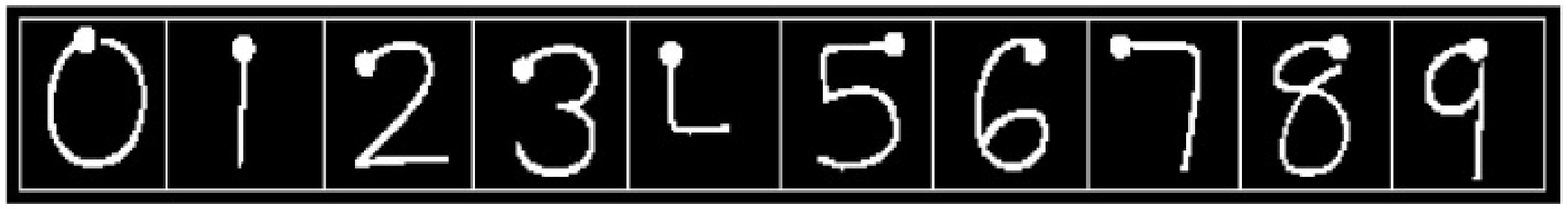}
\caption{Palm's Graffiti digits \cite{2009_Alon}. The dot point indicates the starting position}
\label{Fig:Graffiti}
\end{figure}

\item {\bf In-house dataset}: One limited in-house dataset \cite{2018_Sarma} has been created with the help of three subjects. Here also the gestures are ten Hindu-Western Arabic numerals, 0-9, but in an isolated mode (320$\times$240, 30 frames/s). Dataset consists of 90 videos with 10 classes (three subjects $\times$ ten digits $\times$ three examples/digit/subject). For simplicity, a very simple black background is used with full-sleeve attire worn by the subjects and they used open palm while performing the gestures.
\end{enumerate}


\vspace{-0.1cm}
\subsection{Experimental set-up}
This section gives an idea of the experimental set-up, work performed and the analysis done on the databases to obtain the results. Also, it throws light into the importance of data augmentation as well as a transfer learning process for small databases. The experimentation part is done taking the help of the Google Colab GPU, especially, for the training phase. Other parts of the experiment, is performed in a workstation with Intel\textsuperscript{\textregistered} Core\textsuperscript{TM}i5-4570 CPU at 3.2 GHz and 8GB in RAM without any GPU usage. 

\begin{figure}[htbp]
\centering
\includegraphics[height=0.40\linewidth, width=0.95\linewidth]{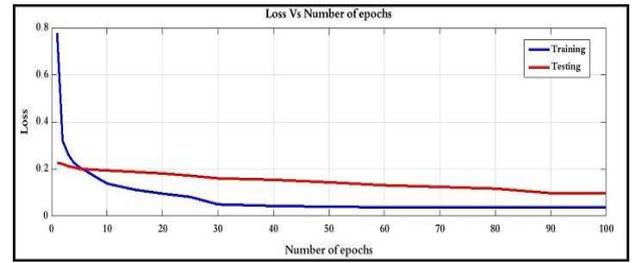}
\caption{Training and testing loss as a function of the number of epochs for Graffiti dataset}
\vspace*{-2mm}
\label{Fig.loss_graph}
\end{figure}

\begin{figure}[h]
\centering
\includegraphics[height=0.40\linewidth, width=0.95\linewidth]{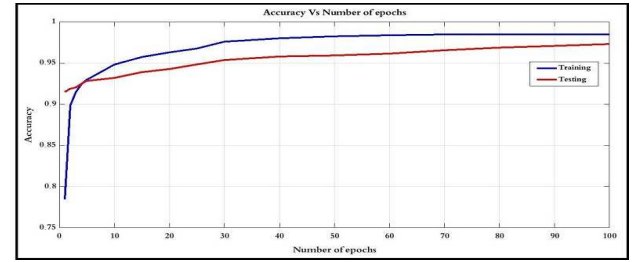}
\caption{Training and testing accuracy as a function of the number of epochs for Graffiti dataset}
\vspace*{-2mm}
\label{Fig.accuracy_graph}
\end{figure}

For training and testing the 3D-CNN model, the segmented video clips of isolated gestures from GreenDigits and EasyDigits sets are used from the Graffiti dataset in a stratified 10-fold cross-validation format. Here stochastic gradient descent (SGD) algorithm is used with the cross-entropy loss function given by Eq. (\ref{eq_22}). The batch size is set to 10 and the model is trained with 100 epochs on individual training datasets. The choice of learning rate is 0.01 if the epoch count is less than 25 and is reduced to 0.001 for epoch count up to 50. To further promote slow learning, values of $1e^{-4}$ and $1e^{-5}$ are used, till 75 and 100 epoch respectively. Fig. (\ref{Fig.loss_graph}) and Fig. (\ref{Fig.accuracy_graph}) gives the training-testing loss and accuracy curves for the Graffiti dataset. From the training loss and accuracy curves, it can be concluded that the system is not suffering from over-fitting after some tuning of the hyper-parameters. Since the gestures of our limited in-house dataset are the same as the training dataset, hence our in-house dataset is used only for testing using the transfer learning process which reduces the burden of training requirements again and again.

\begin{equation} \label{eq_22}
Loss(y,\hat{y}) = -\sum_{j=1}^{M}\sum_{i=1}^{N}y_{ij}log\hat{y_{ij}} 
\end{equation}

\noindent where $M$ is the number of samples, $N$ is the number of classes and $\hat{y}$ is the predicted value for a true value $y$. 

With regard to the training and testing of the 2D-CNN model, the same GreenDigits and EasyDigits sets are used. Since our in-house dataset is not enough for proper training, so data augmentation methods like different affine transformations are used to increase the diversity of data available for training the model, without actually collecting new data. The generation of new data provides robustness as well as scale, translation, and rotation invariance to the system. For data augmentation, several transforms are used like rotation up to 20 degrees, width shift (up to 0.2 range), height shift (up to 0.2 range), sheer (up to 0.2 range), zoom mode (up to 0.2), fill mode on nearest data \textit{etc}. Data augmentation techniques like horizontal and vertical flipping are not used on the images as it may lead to confusion between a few pairs of digits like (2, 5), (4, 7), (6, 9) \textit{etc}. SGD algorithm is carried out with a cross-entropy loss function in the training process. The initial learning rate is set to 0.01 if the epoch count is less than 25 and is reduced to 0.001 for epoch count up to 50 with batch size as 32. The training process is stopped after 50 epochs. 

\subsection{Results}
In this subsection, the performance of the proposed two-stream network is evaluated in three aspects on Graffiti as well as in-house datasets: 2D-CNN model alone with OFMT motion templates as input, 3D-CNN model alone with RGB gesture videos as input, and the combined fusion model. Basically here accuracy, which indicates the proportion of correctly classified samples with respect to the total number of samples, is used as the evaluation index.

In the proposed model, used 3D-CNN and the 2D motion template CNN architecture can be regarded as two heterogeneous networks that can be used as independent models. So, first, we have evaluated the performance of the independent streams and then the fusion performance as a combined network. In the case of 2D-CNN, confusion occurs mainly in discriminating class `3' with class `5' and class `1' with class `7' due to the similarities in their shapes in the OFMT images which can be seen in Fig. (\ref{Fig:OF_MT_images}). Here 3.4\% of `5' are misclassified as `3' or vice-versa and 2.8\% of `7' are misclassified as `1' or vice-versa and rest 2.2\% are various misclassification. Whereas, 3D-CNN has performed quite better in this regard since it also considers the temporal evaluation of the gestures in the video clips. Table (\ref{tab:2DCNN_motion_template}) gives the results of the proposed 2D motion template CNN model, where two cases, without data augmentation and with data augmentation are considered. From Table (\ref{tab:2DCNN_motion_template}), we can conclude that the data augmentation has a great impact on accuracy and can improve the network performance up to a great extent.

\vspace{-0.09cm}
\begin{table}[htbp]
\centering
\caption{Performance accuracy (\%) of 2D-CNN motion template network alone}
\label{tab:2DCNN_motion_template}
\begin{tabular}{ |p{1.3cm}|p{2.9cm}|p{2.9cm}|}
\hline
 Dataset  & Without data augmentation & With data augmentation
      \\ \hline
Graffiti   & 86.24\%      & 92.60\%     \\ \hline
In-house   & 81.20\%      & 89.70\%       \\ \hline
\end{tabular}
\end{table}
To analyze the performance of the 3D-CNN model, stratified 10-fold cross-validation is carried out on combined GreenDigits and EasyDigits sets of Graffiti dataset. The mean accuracy for this dataset is obtained at 97.30\%. Whereas our in-house dataset has provided an accuracy of 96.40\% when tested on the pre-trained network obtained via transfer-learning. Lastly, the prediction class/label scores from both the streams are fused for each class. Since both the streams acquire complementary motion information regarding the gesture, so such fusion generally boosts the recognition performance. That is, the streams complement each-other in acquiring the spatio-temporal information from their respective inputs, and thus certain good output score w.r.t. the target can be achieved, at least by one or the other or by both. In our case also, the same scenario is noticed when both the streams are fused in the decision level. Here decision level fusion is chosen since we have used two non-identical networks applied on inputs with different dimensions. Moreover, rather than simple averaging of the prediction class scores, we have formulated a probabilistic ensemble formula given by Eq. (\ref{eq_20}) with ($\gamma$, $\delta$) as fusion parameters. Different values of $\gamma$ and $\delta$ have been tried and the combination of ($\gamma$, $\delta$) as (0.6, 0.4) provided us the best fusion performance accuracy of 99.20\%. So, here more weight is given to the score provided by the 3D-CNN since it can capture more subtle spatio-temporal features compared to the other one. This is quite justified from the performance achieved by the individual networks performed on the two databases. Our in-house dataset has achieved a recognition rate of 99\% when tested on the fused model.
       
\subsection{Comparison with state-of-the-art}
Our proposed model is compared with three existing methods performed on the same Graffiti dataset. Table (\ref{tab:Graffiti_database_comparison}) represents a comparison of performance for the different methods. The first two methods \cite{2013_Frolova, 2015_Poularakis} rely on hand-crafted feature representations for gesture classification, while \cite{2017_Yang_continuous} uses CNN to extract gesture features. In \cite{2013_Frolova}, the most probable longest common subsequence (MPLCS) is proposed to measure the similarity between the probabilistic template and hand gesture sample. The final decision is based on the probability and length of the extracted subsequences. The method is also compared with HMM and CRF classifiers for performance analysis. Whereas maximum cosine similarity and fastNN is used as a trajectory mapping scheme for digit hand gesture recognition in \cite{2015_Poularakis}. A combined fusion-based method with CNN as trajectory shape recognition and CRF as temporal feature recognition is proposed by \cite{2017_Yang_continuous}. From Table (\ref{tab:Graffiti_database_comparison}), it can be noticed that our 2D-CNN model is as efficient as the classic HMM model, whereas 3D-CNN has achieved even better results. In \cite{2017_Yang_continuous}, fusion-based model with CNN and CRF as components has achieved 98.4\% accuracy which is similar to the sequential state-space MPLCS \cite{2013_Frolova} method. On the other hand, our 2D-CNN and 3D-CNN based fusion model has achieved state-of-the-art results with 99.20\% of accuracy. The fusion result at the decision level has outperformed all other methods, which shows the effectiveness of the fusion scheme. The only work done on our in-house dataset is \cite{2018_Sarma}, which has a similar accuracy of 99\% as this work for alphabet gesture recognition. 
\vspace{-0.2cm}
\begin{table}[htbp]
  \centering
  \caption{Comparison with other methods for pre-segmented Graffiti database}
\begin{tabular}{ |p{0.9cm}|p{1.4cm}|p{1.6cm}|p{1.7cm}|p{1.1cm}|  }
 \hline
 Paper & Feature-type & Features & Classifier & Accuracy\\
 \hline
 \cite{2013_Frolova} & Hand-crafted & Longest common subsequence (LCS)  & HMM, CRF, Most probable LCS (MPLCS) & 89.50\%, 96.40\%, 98.30\% \\
  \hline
 \cite{2015_Poularakis} & Hand-crafted  & Trajectory matching & Max cosine similarity, fastNN & 97.60\% \\
  \hline
  \cite{2017_Yang_continuous} & Both hand-crafted and deep features & CRF-based temporal features & CNN and CRF combined & 98.40\% \\
  \hline
 Our method & Deep features & Deep network, Motion template & only 2D-CNN, only 3D-CNN, Late fusion & 92.60\%, 97.30\%, 99.20\% \\
 \hline
\end{tabular}
\vspace{-0.5cm}
  \label{tab:Graffiti_database_comparison}
\end{table}
\section{Conclusion}
In this work, we propose a two-stream network with 3D-CNN and 2D-CNN as its two streams/layers for hand gesture and in general action recognition. So, the main objective of the model is to detect and recognize isolated dynamic hand gestures with varying shape, size, and color of the hand. This is possible because of the fact that the system doesn't require the pre-segmentation of the hand portion through various methods like skin-segmentation \textit{etc}. The first stream of the system is a 3D-CNN applied for capturing the spatio-temporal information directly from the RGB gesture videos. The second layer is a 2D-CNN model employed to extract motion-patterns for gesture classification. For this stream, an optical flow guided motion template (OFMT) is used as input where the temporal motion information of a gesture is encoded into a single image which helps to remove irrelevant gesture patterns. Moreover, the proposed OFMT can nominally reduce computational complexity and memory requirement as compared to more complex networks like double 3D-CNN/RNN/LSTM models. So, our proposed model can be used in a resource constraint environment without affecting much to its performance. For improving results, the prediction scores of the 3D-CNN model and the 2D-CNN model are fused. Since both the streams acquire complementary motion information regarding the gesture, so such fusion generally boost the recognition performance. Though our model is simple, experimental results have demonstrated that it is able to achieve state-of-the-art results. However, the adopted motion template has the limitation that the moving body has to be in a plane perpendicular to the camera. In the future, we will investigate more on robust feature learning methods for distinguishing the subtle differences among some gesture classes for the viewpoint-invariant gesture recognition method.

\bibliographystyle{IEEEtran} 
\bibliography{DS_OFMT_C3D}

\end{document}